\documentclass[10pt,twocolumn,letterpaper]{article}

\usepackage{cvpr}              

\definecolor{cvprblue}{rgb}{0.21,0.49,0.74}
\usepackage[pagebackref,breaklinks,colorlinks,allcolors=cvprblue]{hyperref}
\usepackage{booktabs}
\usepackage{multirow}
\usepackage{siunitx}
\usepackage{adjustbox}
\usepackage{algorithm}
\usepackage{algorithmic}
\usepackage{caption}
\usepackage{amsmath}

\def\paperID{41615} 
\def\confName{CVPR}
\def\confYear{2026}

\title{Uni-Animator: Towards Unified Visual Colorization}

\author{
    Xinyuan Chen\textsuperscript{1}\thanks{Equal contribution.} \quad
    Yao Xu\textsuperscript{2}\footnotemark[1] \quad
    Shaowen Wang\textsuperscript{1}\footnotemark[1] \quad
    Pengjie Song\textsuperscript{3} \quad
    Bowen Deng\textsuperscript{4} \\[1ex]
    \textsuperscript{1}Mississippi State University \quad
    \textsuperscript{2}UIUC \quad
    \textsuperscript{3}Hunan University \quad
    \textsuperscript{4}HKUST \\[1ex]
    {\tt\small email@domain}
}

\begin{document}

\maketitle

\begin{abstract}

We propose Uni-Animator, a novel Diffusion Transformer (DiT)-based framework for unified image and video sketch colorization. Existing sketch colorization methods struggle to unify image and video tasks, suffering from imprecise color transfer with single or multiple references, inadequate preservation of high-frequency physical details, and compromised temporal coherence with motion artifacts in large-motion scenes. To tackle imprecise color transfer, we introduce visual reference enhancement via instance patch embedding, enabling precise alignment and fusion of reference color information. To resolve insufficient physical detail preservation, we design physical detail reinforcement using physical features that effectively capture and retain high-frequency textures. To mitigate motion-induced temporal inconsistency, we propose sketch-based dynamic RoPE encoding that adaptively models motion-aware spatial-temporal dependencies. Extensive experimental results demonstrate that Uni-Animator achieves competitive performance on both image and video sketch colorization, matching that of task-specific methods while unlocking unified cross-domain capabilities with high detail fidelity and robust temporal consistency.

\end{abstract}    
\section{Introduction}
\label{sec:intro}
Sketch colorization is a foundational task in digital content creation, with pivotal applications in animation production, film restoration, and game development. Traditional workflows are predominantly manual: static sketches require artists to meticulously layer colors, balance style consistency, and refine textures, an expertise-dependent and time-consuming process, while video colorization demands frame-by-frame adjustments to maintain temporal coherence and suppress flickering, further exacerbating labor costs. Despite delivering high-quality outputs, manual methods suffer from inherent bottlenecks: prohibitive scalability for long sequences, low production efficiency, and misalignment with modern industrial demands. These limitations drive the need for automated solutions that preserve artistic fidelity while unifying accessibility across both image and video domains.

\begin{figure}[t]
    \centering\includegraphics[width=\linewidth]{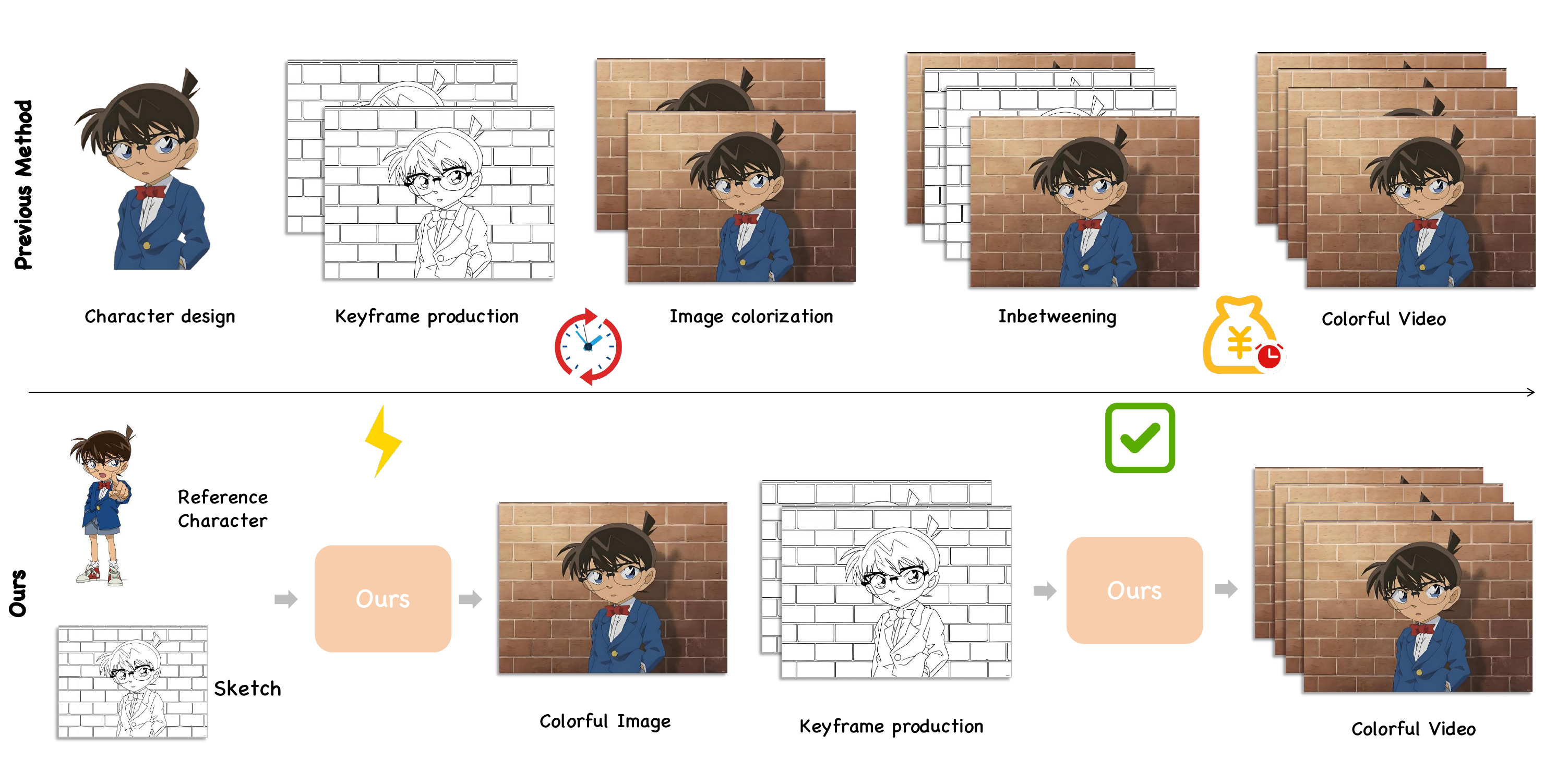}
    \caption{\textbf{Workflow Comparison: Manual Colorization vs. Ours}. Our automated framework eliminates repetitive frame-by-frame adjustments and cross-domain adaptation efforts, significantly reducing labor costs in industrial production.}
    \label{fig: workflow}
\end{figure}

\begin{figure}[t]
    \centering
    \includegraphics[width=\linewidth]{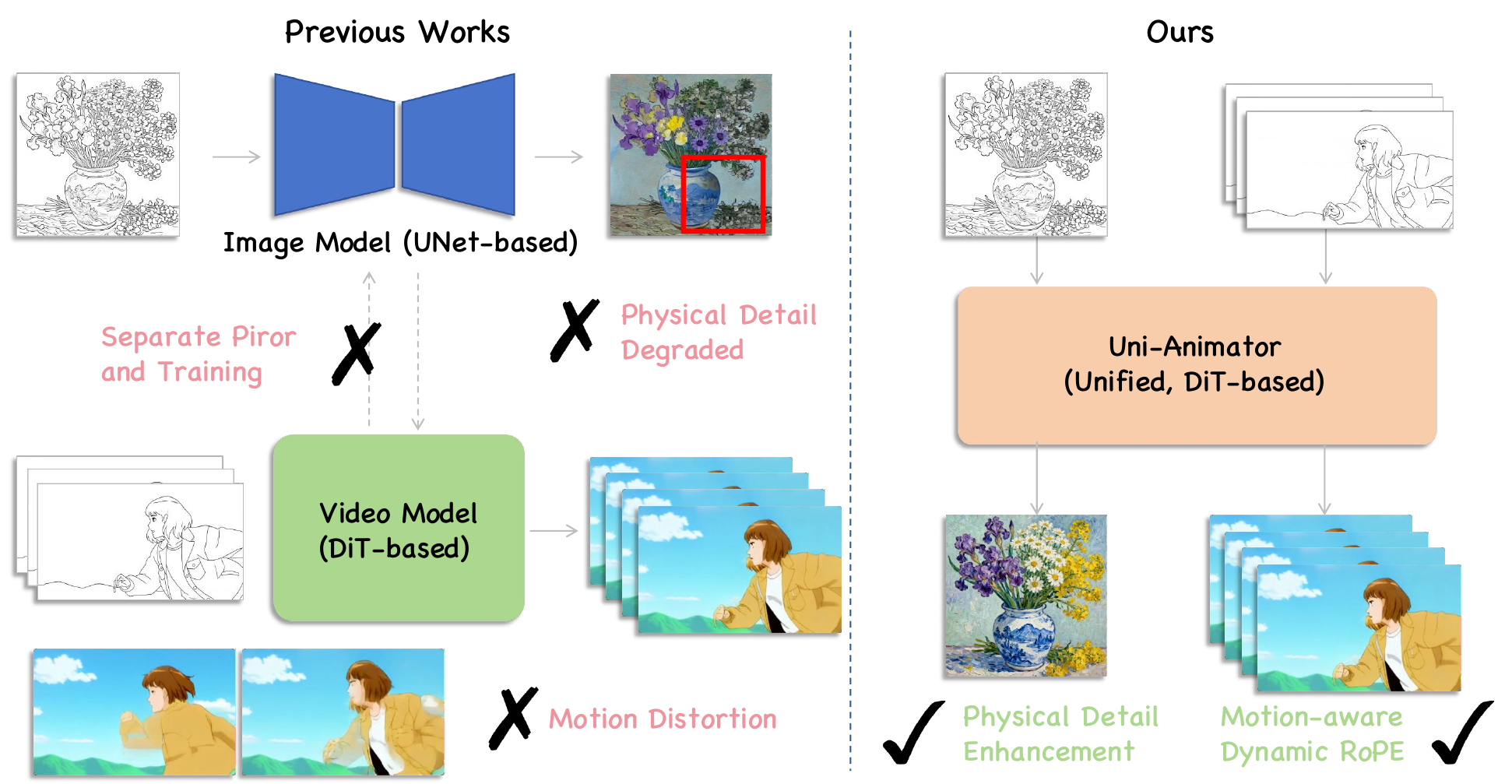}
    \caption{\textbf{Motivation: Limitations of Existing Methods vs. Our Solution}. Existing methods suffer from style deviation (left), detail degradation (right), and temporal flickering (bottom) when handling images or videos separately. Our unified framework addresses all three issues concurrently.}
    \label{fig: motivation}
\end{figure}

The rising demand for high-fidelity sketch colorization has spurred advancements in automated methods, particularly diffusion models \cite{sketch-to-photo, coloridiffusionv2, z1, z2, z3, z4, zhang2025easycontrol, zhang2024ssr, song2025layertracer, song2025makeanything, huang2025photodoodle, Cobra}. However, existing approaches still face three unresolved challenges that hinder practical deployment, compounded by a lack of unified frameworks for cross-domain (image/video) processing:
Existing methods are inherently domain-specialized, either optimized for static images \cite{liu2025manganinja, followyourcolor} or dynamic videos \cite{meng2024anidoc, tooncomposer}, failing to adapt to mixed-content production pipelines. Beyond domain separation, three core technical limitations persist:
\textbf{(1) Inadequate Visual Reference Utilization.} Diffusion-based methods often fail to fully exploit reference images, as global feature extraction neglects local color distributions, textures, and lighting details. This leads to outputs that deviate from the target artistic style and lack fine-grained consistency with references \cite{AnimeColor, huang2024lvcd}.
\textbf{(2) High-Frequency Detail Degradation.} Variational Autoencoders (VAEs) in diffusion pipelines introduce irreversible compression artifacts, erasing critical high-frequency details (e.g., metallic reflections, texture granularity) and compromising physical consistency. Methods like ToonComposer \cite{tooncomposer} struggle to recover these details, limiting applicability to high-precision production \cite{xing2024tooncrafter}.
\textbf{(3) Temporal Inconsistency in Dynamic Scenes.} Video-specialized methods lack effective motion modeling for sketch sequences, resulting in flickering and misalignment during non-uniform movements (e.g., sudden character motions). As shown in Figure~\ref{fig: motivation}, inadequate temporal modeling fails to bind color and motion across frames \cite{LayerAnimate, Magic3DSketch}.

To bridge these gaps, we propose \textbf{Uni-Animator}, a unified diffusion-based framework for high-fidelity sketch colorization that seamlessly supports both image and video inputs. Our framework targets the three core limitations above with coordinated technical innovations:
\textbf{Visual Reference Enhancement}: To address inadequate reference utilization, we design a \textit{Visual Reference Enhancement} mechanism. It consists of an \textit{Instance Patch Embedding} module that extracts localized fine-grained features from references by partitioning images into patches and performing independent feature encoding, capturing local color distributions and textures more precisely than global extraction. We further inject VAE-encoded reference features directly into the noisy latent space, enabling semantic alignment between references and input sketches during denoising.
\textbf{Physical Detail Reinforcement}: To mitigate VAE-induced detail loss, we leverage pre-trained DINO models to extract rich high-frequency features (encoding material properties, surface textures, and lighting). These features are concatenated along the token dimension and fused into the DiT backbone via in-context learning, enabling the model to preserve edge sharpness and physical details while maintaining color consistency.
\textbf{Sketch-Based Dynamic RoPE}: For temporal coherence in dynamic scenes, we propose \textit{Sketch-Based Dynamic RoPE}, a motion-aware positional encoding strategy. We extract optical flow from sketch sequences, decompose it into horizontal and vertical motion components, and dynamically adjust RoPE frequencies based on motion intensity: higher frequencies for fast-moving regions (to capture dynamic details) and default frequencies for static regions (to maintain stability). This design specifically targets sketch motion characteristics, suppressing flickering and misalignment.
Our core contributions are summarized as follows:
\begin{itemize}
    \item We propose \textbf{Uni-Animator}, the first unified framework that supports high-fidelity sketch colorization for both images and videos, eliminating the need for task-specific model adaptation.
    \item Technically, we design Visual Reference Enhancement to precisely capture the style and Physical Detail Reinforcement to tackle inadequate visual and physical degradation. Sketch-Based Dynamic RoPE suppresses flickering and misalignment in dynamic scenes to improve temporal consistency.
    \item Our framework achieves state-of-the-art performance in both image and video colorization, with superior reference fidelity, detail preservation, and temporal coherence.
\end{itemize}

\section{Related Work}

\subsection{Image line drawing Colorization}
Image line art colorization has evolved from manual-dependent traditional methods to data-driven deep learning paradigms, with key advancements addressing generalization and style controllability.
Early traditional methods relied on either optimization-based color propagation or graph cut algorithms. Optimization approaches leveraged color continuity assumptions (adjacent pixels share hues) and manual annotations to propagate colors via heuristic rules, whereas graph-cut methods framed color filling as a graph-theoretic minimum-cut problem. Both suffered from heavy manual workloads and poor adaptability to complex line art structures (e.g., intricate strokes), limiting real-world utility.
Deep learning shifted the paradigm to data-driven solutions, with GANs \cite{style2paint, CGAN, sketch-to-photo, ma2024followpose, ma2025followcreation, ma2026fastvmt, ma2025followyourmotion, ma2025followfaster, ma2025controllable, controlnet, AnimeDiffusion, SGA, BigColor,wang2024taming,feng2025dit4edit,wang2024cove,chen2025infinite,zhu2025multibooth,zhu2024instantswap,yan2025eedit,zhang2025magiccolor,liu2025avatarartist,liu2025multimotion, ChromaGAN, pix2pix} becoming the first mainstream approach. Through adversarial training, where a generator maps line art to colorized images and a discriminator distinguishes real vs. synthetic outputs, GANs modeled realistic color distributions. However, inherent mode collapse restricted style diversity, producing homogeneous results.
Recent diffusion models overcame this limitation, offering enhanced controllability and stability. Representative works like MangaNinja \cite{liu2025manganinja, coloridiffusionv2, ColorFlow, Cobra, followyourcolor, ye2023ip} integrated reference images via dedicated fusion modules and point-driven control schemes for fine-grained color matching. Despite progress in style consistency, diffusion models still struggle with high-frequency detail preservation (e.g., sharp edges, fine textures) due to VAE compression or insufficient local feature modeling.

\subsection{Video Sketch Colorization}

Video sketch colorization builds on image colorization but adds the critical constraint of temporal coherence, driving a separate evolution focused on balancing frame quality and inter-frame stability.
Early methods extended GANs or variants \cite{VCGAN,blattmann2023stable, Magic3DSketch, xing2024make, su2024roformer, loshchilov2017decoupled, videointerpolation, ho2022video, blattmann2023align, jiang2024exploring, Dynamicrafter, zhu2024champ, hu2024animate} to process frames independently, using temporal loss terms (e.g., optical flow-based consistency) to suppress flickering. While achieving basic stability, they lacked explicit temporal modeling, leading to cumulative inconsistencies in long sequences and persistent mode collapse from GANs.
The shift to diffusion models addressed these gaps by leveraging their superior generative quality and compatibility with temporal modeling. Early diffusion-based video methods like Anidoc \cite{meng2024anidoc, LayerAnimate, coloridiffusionv2, Cobra, ma2024followyouremoji, ma2025followyourclick, ColorFlow,followyourcolor, videointerpolation, ma2024follow} extended image diffusion frameworks by modeling the denoising process across the temporal dimension, enabling coherent video generation from line art sequences. However, they still suffered from high-frequency detail loss. VAE-induced compression erased fine textures and edge sharpness, leading to blurry outputs in detail-rich regions (e.g., metallic surfaces or fabric textures).
To further enhance temporal coherence, recent works \cite{tooncomposer, AnimeColor, huang2024lvcd, tooncomposer} integrated Diffusion Transformers (DiT) \cite{dit, yang2024cogvideox, long2025follow, ho2020denoising, LayerT2V, LayerAnimate, kong2024hunyuanvideo, wang2025wan, wang2025cinemaster} as backbones, leveraging self-attention to capture long-range spatiotemporal dependencies. This design improved inter-frame consistency by modeling global temporal context, outperforming UNet-based methods in handling slow-moving scenes. Nevertheless, DiT-based approaches still struggle with fast-moving objects or non-uniform motions: the self-attention mechanism’s global information aggregation leads to misalignment when motion intensity is high, resulting in motion artifacts and flickering. Critically, all existing video methods remain domain-specialized, but they cannot generalize to static image inputs, creating a gap in mixed-content production pipelines.

\begin{figure*}[t]
    \centering
    \includegraphics[width=\linewidth]{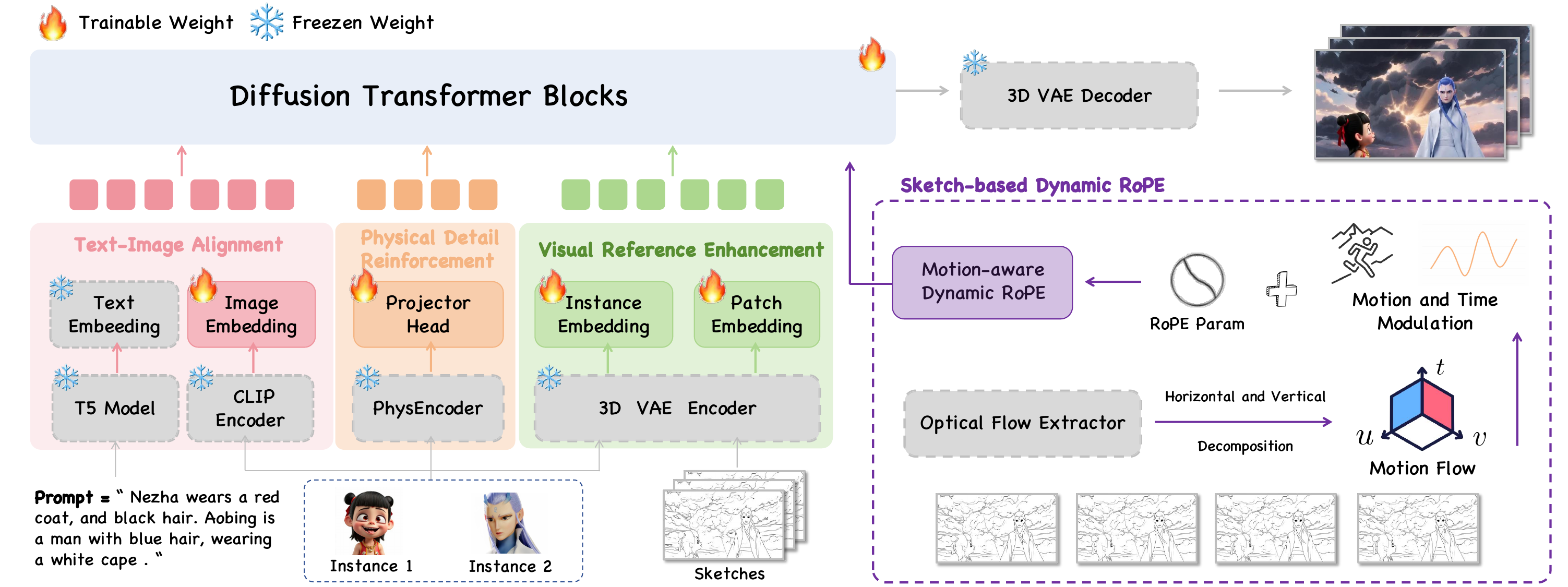}
    \caption{\textbf{Overall architecture of Uni-Animator.} Given input visual references, the VAE encoder and physical projector encode its physical features and concatenate with visual latent features, then the DiT model predicts subsequent frames conditioned on physical, visual, and text embeddings. Sketch-based Dynamic RoPE to improve
    temporal inconsistency in large motion dynamic scenes without extra training.}
    \label{fig: framework}
\end{figure*}

\section{Method}

We introduce Uni-Animator, a DiT Transformer that ingests text, sketches, and visual references and produces colorful image sequences. We start by introducing the problem and architecture in Sec .~\ref {sec: problem}. The Visual Reference Fusion in Sec .~\ref {sec: visual_ref}, Physical Detail Reinforcement in Sec .~\ref {sec: physic}, and finally the Sketch-based Dynamic RoPE in Sec .~\ref {sec: rope}.

\subsection{Problem Definition and Architecture}
\label{sec: problem}

We formalize the sketch-guided unified image and video colorization task as follows: Given a sketch frame sequence \( S = \{s_1, ..., s_M\} \) (where \( M \) denotes the number of sketch frames, with \( M=1 \) corresponding to the image colorization task), a text prompt $\text{txt}$, and a set of visual references $\text{Ref} = \{r_1, ..., r_N\} $ (where \( N \) is the number of visual references), our framework aims to generate a sequence of colorful frames \( V_{\text{gen}} = \{v_1, ..., v_M\} \) that strictly aligns with the structure of \( S \) and the semantics of \( txt \), while preserving fine-grained details. The overall task is concisely formulated as:
\begin{equation}
    V_{\text{gen}} = \mathbf{F}(S, \text{txt}, \text{Ref}),
\end{equation}
where \( \mathbf{F} \) denotes our proposed Uni-Animator framework. The training objective is formularize as follows:

\begin{equation}
\mathcal{L}=\mathbb{E}_{z_{0}, t, c_{\text{txt}}, c_{\text{skt}}, \epsilon} \left[\left\| \epsilon-\epsilon_{\theta}\left( z_{t}, c_{\text{txt}}, c_{\text{skt}}, t\right) \right\|_{2}^{2} \right],
\tag{5}
\end{equation}
where $\epsilon \sim \mathcal{N}(0,1)$ is randomly sampled from standard Gaussian noise, $t \in 1, \ldots, T$ denotes the diffusion timestep, $z_0$ represents the video latent encoded by VAE encoder, $z_t$ is the noised latent.

\paragraph{Architecture:} Uni-Animator is implemented upon a diffusion transformer model (DiT) initialized from Wan2.1~\cite{wan}, which employs a 3D Variational Autoencoder (VAE) to transform videos and the initial frame to latent space, CLIP encoder for image embedding, and a T5 encoder~\cite{t5} for text embeddings. It is worth expecting that the learned high-frequency representation can be used as a generalizable guidance of both physical properties and dynamics for physics-aware image and video generation. We built PhysEncoder with a DINOv2~\cite{DINOv2}. The former adopts pretrained weights for initialization and takes the role of semantic perception, while the latter adapts the extracted high-level semantic features into an appropriate dimension to be injected into the DiT model. 

\begin{figure}[t]
    \centering
    \includegraphics[width=\linewidth]{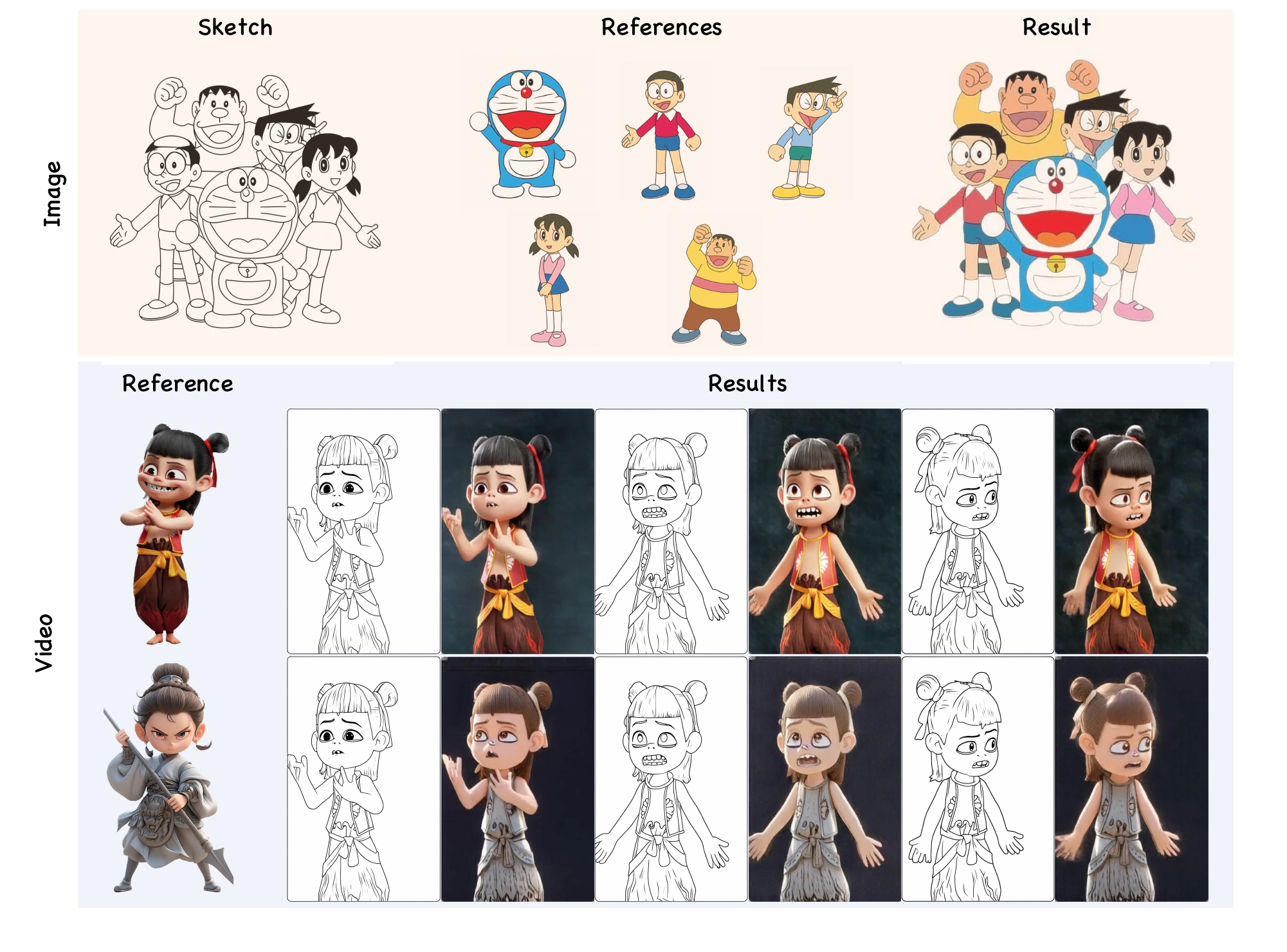}
   \caption{\textbf{Visual Control Ability}. Our method supports accurate semantic matching for image colorization with over five visual references. For video colorization, it enables dynamic customization of characters by altering visual references while preserving original motions.}
    \label{fig: framework}
\end{figure}

\subsection{Visual Reference Fusion}
\label{sec: visual_ref}
The core goal of visual reference fusion is to fully leverage the semantic and structural information from reference images, enabling the DiT to explicitly model inter-frame visual correspondences while supplementing fine-grained details thus ensuring both temporal consistency and detail fidelity in the colorization results. To this end, we design a two-stage multi-granularity fusion strategy that integrates reference information into the diffusion process without disrupting the noise prediction pipeline, combining channel-wise coarse-grained feature fusion and sequence-wise fine-grained feature fusion. 
For coarse-grained fusion, which aims to inject global structural and style information from reference images into the initial noise latent space, we first let \( \mathcal{R} = \{R_1, R_2, ..., R_n\} \) denote a sequence of reference images, where \( N \) is the number of reference frames and \( R_n \in \mathbb{R}^{H \times W \times 3} \) for \( n=1,..., N \)) and employ a pre-trained Variational Autoencoder (VAE) encoder \( \mathcal{E}(\cdot) \) to map \( \mathcal{R} \) into a low-dimensional coarse-grained latent space as
$
\mathcal{I} = \mathcal{E}(\mathcal{R}) \in \mathbb{R}^{b \times c \times T \times h \times w},
$
where \( b \) as batch size, \( c \) as latent channel dimension, \( h = H/s \), \( w = W/s \), and \( s \) as VAE downsampling factor and \( \mathcal{I} \) encodes the global style, structure, and temporal information of the reference sequence. We then initialize the noise latent \( \mathbf{z}_0 \sim \mathcal{N}(0, \mathbf{I}) \in \mathbb{R}^{b \times c \times T \times h \times w} \) and fuse \( \mathbf{z}_0 \) and \( \mathcal{I} \) along the channel dimension to get the coarse-grained latent as
\begin{equation}
     \mathbf{z}_{\text{coarse}} = \text{Concat}(\mathbf{z}_0, \mathcal{I}) \in \mathbb{R}^{b \times (C + N) \times T \times h \times w} ;
\end{equation}
While coarse-grained fusion captures global structure, it cannot model local instance-level details (e.g., texture patterns, edge semantics), so we design a sequence-wise fine-grained fusion strategy to address this gap: we first convert \( \mathbf{z}_{\text{coarse}} \) into a sequence of patch features using a patching embedding layer, which splits the spatial dimensions \( (h, w) \) into non-overlapping \( p \times p \) patches and projects each patch into a high-dimensional space to obtain 

\begin{equation}
    \mathbf{z}_{\text{noise}} = \text{PatchEmbedding}(\mathbf{z}_{\text{coarse}}) \in \mathbb{R}^{b \times Seq \times d},
\end{equation}
with \( Seq \) as total sequence length and \( d \) as embedding dimension, then apply a convolutional instance embedding module \( \mathcal{I}_\text{emb}(\cdot) \) with stacked 3D convolutions to capture local spatial-temporal correlations to \( \mathcal{R} \) for extracting fine-grained instance-level features as 
\begin{equation}
    \mathbf{z}_{\text{ref}} = \mathcal{I}_\text{emb}(\mathcal{R}) \in \mathbb{R}^{b \times N \times d},
\end{equation}
And finally fuse these two features along the sequence dimension to get \( \mathbf{z}_{\text{fused}} = \text{Concat}(\mathbf{z}_{\text{noise}}, \mathbf{z}_{\text{ref}}) \in \mathbb{R}^{b \times (N + Seq) \times d} \). This two-stage fusion integrates complementary information at different granularities—coarse-grained channel fusion provides global style and inter-frame correspondence guidance to ensure temporal consistency and style alignment, while fine-grained sequence fusion supplements instance-level texture and detail information to enhance local structure fidelity and feeding \( \mathbf{z}_{\text{fused}} \) into DiT enables the model to simultaneously exploit global inter-frame correlations and local instance details from references, laying the foundation for high-quality, consistent sketch colorization.

\subsection{Physical Detail Reinforcement}
\label{sec: physic}
To further enhance the performance of sketch colorization specifically addressing the lack of physical structural fidelity and inter-modal semantic conflicts in prior fusion strategies this module focuses on two core objectives: integrating discriminative physical representations to reinforce structural details and resolving semantic inconsistencies between multi-source visual references and text descriptions (e.g., cases where the text specifies "pink hair" but visual references depict hair of other colors, leading to misaligned colorization results). 

For physical representation reinforcement, we leverage DINOv2, a pre-trained vision transformer renowned for extracting robust, structure-aware visual features, to independently encode the visual reference sequence \( \mathcal{R} \), yielding a physical feature embedding \( \mathbf{z}_{\text{physic}} \in \mathbb{R}^{b \times N \times d_{\text{physic}}} \) where \( P \) denotes the number of patches output by DINOv2 and \( d_{\text{physic}} \) is the dimension of DINOv2’s patch features; to align this physical feature with the dimension of the fused latent \( \mathbf{z}_{\text{fused}} \) and enhance its discriminability, we design a physical head consisting of stacked MLP layers that projects \( \mathbf{z}_{\text{physic}} \) into the target embedding space \( d \), resulting in 
\begin{equation}
    \hat{\mathbf{z}}_{\text{physic}} = \text{Head}_{\text{physic}}(\mathbf{z}_{\text{physic}}) \in \mathbb{R}^{b \times P \times d};
\end{equation}
We then concatenate this processed physical feature with \( \mathbf{z}_{\text{fused}} \in \mathbb{R}^{b \times Seq \times d} \) (where \( Seq \) is the sequence length of the fused latent from fine-grained fusion) along the sequence dimension to integrate physical structural information, yielding the enhanced latent feature 
\begin{equation}
    \mathbf{z} = \text{Concat}(\mathbf{z}_{\text{fused}}, \hat{\mathbf{z}}_{\text{physic}}) \in \mathbb{R}^{b \times (Seq + P) \times d}.
\end{equation}

To resolve semantic conflicts between multiple visual references and text descriptions, we draw inspiration from image-to-video alignment methods and adopt CLIP, a pre-trained cross-modal encoder, to model inter-modal consistency: we first encode each visual reference \( R_t \in \mathcal{R} \) via CLIP’s vision encoder to obtain visual embeddings 
\begin{equation}
    \mathbf{c}_{\text{vis}} = \{\text{CLIP}_{\text{vis}}(R_1), ..., \text{CLIP}_{\text{vis}}(R_T)\} \in \mathbb{R}^{b \times N \times d_{\text{clip}}},
\end{equation}
and encode the text prompt via T5 encoder to get the text embedding, then concatenate \( \mathbf{c}_{\text{text}} \) with \( \mathbf{c}_{\text{vis}} \) to form a unified cross-modal condition. Finally, we introduce a cross-attention layer that takes \( \mathbf{z}_{\text{phys-enhanced}} \) as queries, \( \mathbf{c}_{\text{cross}} \) as keys and values, to model dependencies between the latent feature and cross-modal conditions, formulated as 
\begin{equation}
    \mathbf{z}_{\text{final}} = \text{CrossAttn}(\mathbf{z}_{\text{phys-enhanced}}, \mathbf{c}_{\text{text}}, \mathbf{c}_{\text{vis}}),
\end{equation}
which adaptively weights consistent information across modalities and suppresses conflicting signals, thereby mitigating semantic misalignment between text and visual references.

\subsection{Sketch-based Dynamic RoPE}
\label{sec: rope}

The problem of motion flicker between frames remains a significant challenge in video sketch colorization, particularly for sketch sequences with large motion amplitudes where conventional positional encoding strategies often fail to capture rapid dynamics, leading to motion artifacts and flickering. To address this limitation, we propose a Sketch-based Dynamic RoPE mechanism that adaptively adjusts frequency parameters based on motion intensity extracted directly from sketch sequences.

\paragraph{Motion-Aware Frequency Adaptation} Our approach begins by extracting optical flow information from consecutive sketch frames to quantify motion characteristics. As shown in Algorithm~\ref{alg:dynamic_rope}, we employ the RAFT architecture~\cite{RAFT} for robust flow estimation. The extracted flow fields are decomposed into horizontal ($u$) and vertical ($v$) components, from which we compute motion magnitude $M = \sqrt{u^2 + v^2}$.

The core innovation lies in our dimension, specific frequency scaling strategy, where RoPE parameters are dynamically adjusted according to motion characteristics in different spatial-temporal dimensions:
\begin{equation}
f_{\text{dynamic}} = f_{\text{base}} \cdot (1 + \alpha \cdot \hat{M}),
\end{equation}
where $f_{\text{base}}$ represents the original RoPE frequency, $\hat{M}$ is the normalized motion intensity in $[0,1]$, and $\alpha$ is a scaling factor controlling adaptation intensity.

\paragraph{Dimension-Specific Motion Adaptation} We extend this adaptation across three spatial-temporal dimensions with specialized scaling strategies:
\textbf{Temporal Dimension:} Global motion intensity modulates temporal frequencies to enhance inter-frame coherence. The rationale is that higher motion regions require finer temporal resolution to accurately capture rapid changes and prevent temporal aliasing. This ensures smooth color transitions during dynamic actions while maintaining efficiency in static regions.
\textbf{Height Dimension:} 
Vertical motion components ($v$) specifically influence height frequencies. This directional adaptation preserves edge coherence along vertical structures during up-down movements, which is particularly important for maintaining character silhouettes and vertical elements in sketch animation.
\textbf{Width Dimension:} Horizontal motion components ($u$) modulate width frequencies. This adaptation enhances the model's ability to track horizontal displacements, crucial for preserving spatial consistency during lateral movements common in character animation and panning shots.

\begin{algorithm}[t]
\caption{Sketch-based Dynamic RoPE Generation}
\label{alg:dynamic_rope}
\begin{algorithmic}[1]
\REQUIRE Sketch video sequence $V \in \mathbb{R}^{T \times C \times H \times W}$, 
        feature dimension $d$, 
        max sequence length $L$,
        scaling factors $\alpha_t = 0.1, \alpha_h = 0.3, \alpha_w = 0.3$
\ENSURE Dynamic RoPE frequencies $\mathbf{F}_{\text{dynamic}} \in \mathbb{C}^{L \times d/2}$

\STATE Initialize RAFT flow estimator with pre-trained weights
\FOR{each consecutive frame pair $(V_t, V_{t+1}) \in V$}
    \STATE $\mathbf{flow} \leftarrow \text{RAFT}(V_t, V_{t+1})$  
    \STATE $u, v \leftarrow \text{decompose}(\mathbf{flow})$    
    \STATE $M \leftarrow \sqrt{u^2 + v^2}$                     
\ENDFOR

\STATE $\hat{M}_{\text{global}} \leftarrow \text{normalize}(\text{mean}(M))$  
\STATE $\hat{M}_v \leftarrow \text{normalize}(\text{mean}(v))$  \COMMENT{Vertical motion along height}
\STATE $\hat{M}_u \leftarrow \text{normalize}(\text{mean}(u)$  \COMMENT{Horizontal motion along width}

\STATE $d_t, d_h, d_w \leftarrow \text{split\_dimensions}(d)$  

\STATE $\mathbf{F}_{\text{time}} \leftarrow \text{compute\_dynamic\_rope}(L, d_t, \hat{M}_{\text{global}}, \alpha_t)$
\STATE $\mathbf{F}_{\text{height}} \leftarrow \text{compute\_dynamic\_rope}(L, d_h, \hat{M}_v, \alpha_h)$
\STATE $\mathbf{F}_{\text{width}} \leftarrow \text{compute\_dynamic\_rope}(L, d_w, \hat{M}_u, \alpha_w)$

\STATE $\mathbf{F}_{\text{dynamic}} \leftarrow \text{concat}(\mathbf{F}_{\text{time}}, \mathbf{F}_{\text{height}}, \mathbf{F}_{\text{width}})$
\RETURN $\mathbf{F}_{\text{dynamic}}$
\end{algorithmic}
\end{algorithm}

\begin{figure}[t]
    \centering
    \includegraphics[width=0.95\linewidth]{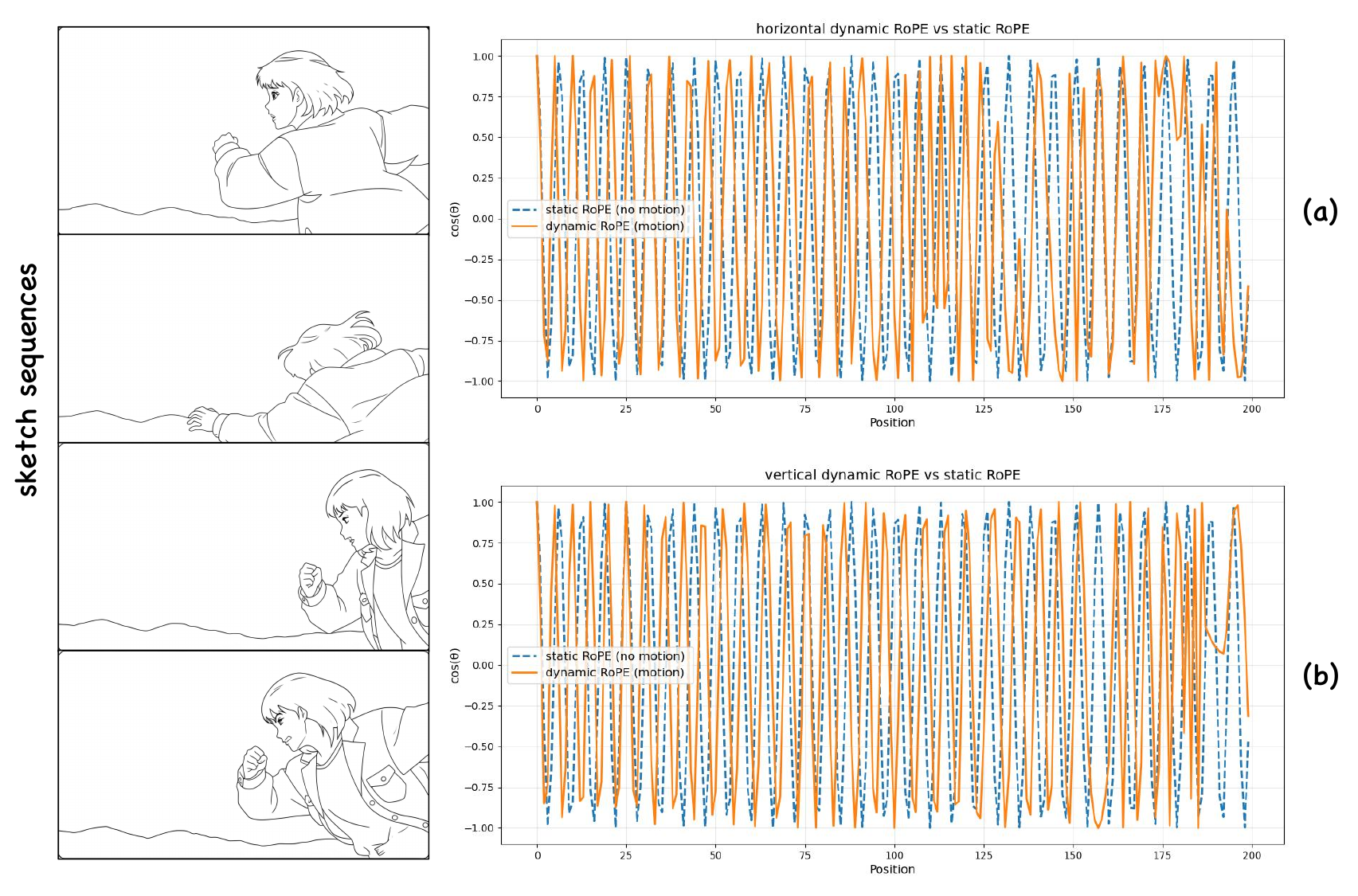}
    \caption{\textbf{Visualization of RoPE frequency comparison.} Strong horizontal motion (a) triggers higher frequency oscillations compared to gentle vertical motion (b), matching the asymmetric motion patterns observed in sketch sequences.}
    \label{fig:rope_visualization}
\end{figure}

The dynamic RoPE is integrated into our transformer architecture by replacing standard positional encodings. 
A Higher value accommodates the need for enhanced temporal resolution during rapid motion, preventing flickering artifacts in high-dynamic scenes.
Conservative values maintain spatial stability while allowing directional motion adaptation, crucial for preserving sketch structural integrity.
For sequences with minimal motion ($\hat{M} < 0.1$), the mechanism gracefully degrades to standard RoPE, ensuring backward compatibility.
As shown in Figure~\ref{fig:rope_visualization}, the comparative analysis reveals that sketch sequences often exhibit asymmetric motion patterns, with intense horizontal movement accompanied by gentle vertical motion. Our dynamic RoPE effectively captures this asymmetry: stronger horizontal motion triggers higher frequency adaptations, manifested as more rapid cosine value oscillations. This frequency escalation ensures optimal encoding resource allocation, prioritizing dimensions with significant motion while maintaining stability in relatively static regions.

\begin{figure*}[t]
\centering
\includegraphics[width=\linewidth]{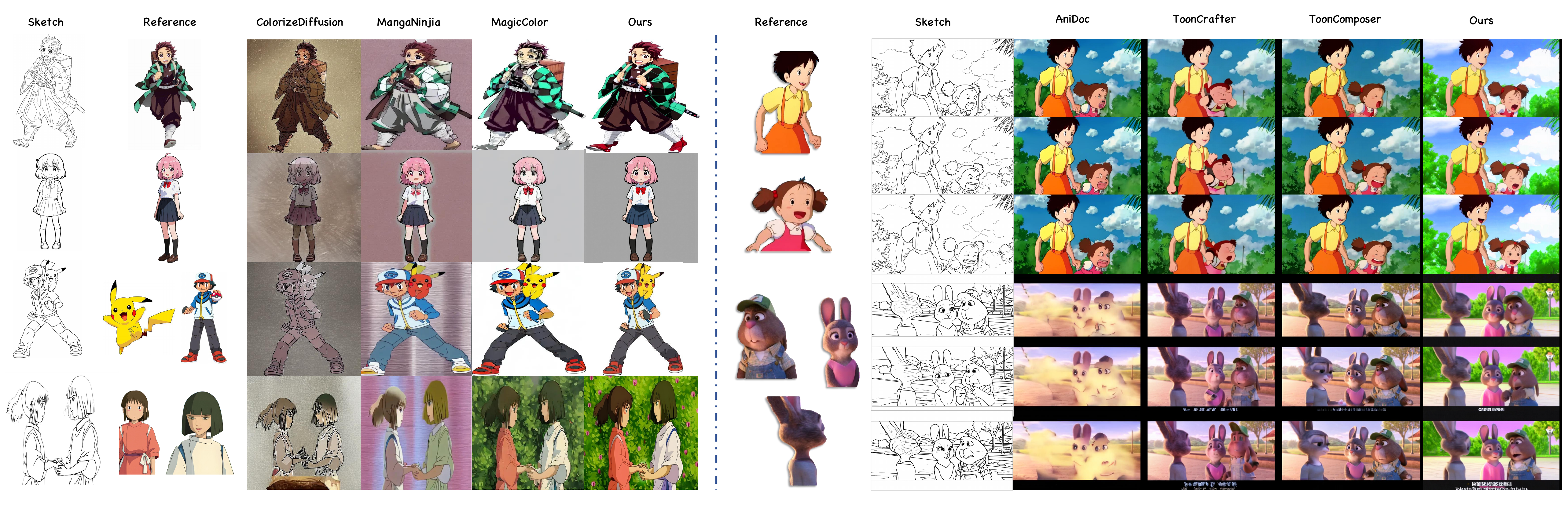}
\caption{\textbf{Comparison of Visual Results with Baselines}. Our method shows better visual consistency during image and video sketch colorization, supporting single to multiple visual references dynamically. Compared with other video methods that use the first frame as input, our method shows a competitive result and the ability to customize the coloring results without the first frame.}
\label{fig: comparison}
\end{figure*}

\section{Experiment}

\subsection{Implementation Details}
\label{sec: implementation}

\textbf{Training Setting.} Uni-Animator is initialized from the pre-trained Wan2.1-14B image-to-video (I2V) model and fine-tuned via Low-Rank Adaptation~\cite{hu2022lora} with rank 64, using a batch size of 1, initial learning rate of \(2 \times 10^{-5}\), AdamW optimizer (weight decay \(3 \times 10^{-2}\)). Images and videos are trained in proportion, with images sampled from the video data. \textbf{Dataset.} Our training dataset combines 5K manually curated anime video clips (from 5 films like \textit{Spirited Away}, \textit{Doraemon}) preprocessed via frame extraction, sketch generation, and quality filtering, and 30K filtered clips from the Sakuga-42M dataset~\cite{sakuga}, all normalized to \(512 \times 512\) single-channel grayscale sketches. The test dataset was randomly sampled from animation films, unlike the train dataset.
\textbf{Evaluation Metric.} We evaluate performance using SSIM~\cite{ssim} and FID (image quality), CLIP Score (style/content alignment), and LIPIS~\cite{lapips} with a custom Temporal Consistency Score (temporal coherence), where lower FID/LIPIS and higher SSIM/CLIP Score/Temporal Consistency Score indicate better performance.

\subsection{Quantitative Results}

To systematically evaluate video and image sketch colorization performance, we constructed a rigorous test set: 100 high-quality animation video clips (with multiple reference instances) and 1.5K sampled image pairs, one for extracting sketch and the other for visual reference. For each video clip, we manually extracted character instances as reference images via SAM \cite{sam}, ensuring consistent evaluation conditions across all baselines. We compare our framework against state-of-the-art image and video colorization methods, adhering to each method’s default input configurations for fairness: image colorization methods receive a single reference image, video colorization methods use the first frame of the original video as reference, while our method uniquely leverages extracted instance-level references for both image and video inputs.
As shown in Table~\ref{tab: comparison}, our method achieves superior overall performance across tasks: it ranks first in video colorization, with FID and LPIPS metrics far outperforming all baselines, a testament to its superior visual fidelity and consistency. While ToonComposer leads in the CLIP metric, we maintain a strong second-place finish; for image colorization, we also secure top performance, with MagicColor only leading in PSNR. Critically, our framework is the only one that delivers excellent results in both image and video tasks, demonstrating its cross-domain adaptability and instance-aware controllability. This dual-domain superiority highlights the practical value of our unified design for mixed-content production pipelines.

\begin{table}[h]
\centering
\caption{Quantitative comparison of different methods on image and video colorization tasks.}
\label{tab: comparison}
\begin{adjustbox}{width=\linewidth}
\begin{tabular}{lcccccc}
\toprule
\textbf{Task} & \textbf{Method} & \textbf{SSIM $\uparrow$} & \textbf{PSNR $\uparrow$} & \textbf{CLIP $\uparrow$} & \textbf{FID $\downarrow$} & \textbf{LPIPS $\downarrow$} \\
\midrule
\multirow{4}{*}{Image Colorization} 
& ColorizeDiffusion~\cite{coloridiffusionv2} & 0.512 & 17.882 & 0.762 & 82.201 & 0.557 \\
& MangaNinja~\cite{liu2025manganinja} & 0.543 & 14.289 & 0.728 & 43.165 & 0.425 \\
& MagicColor~\cite{followyourcolor} & 0.713 & \textbf{23.751} & 0.798 & 28.953 & 0.231 \\
& \textbf{Ours} & \textbf{0.741} & 21.961 & \textbf{0.845} & \textbf{23.761} & \textbf{0.211} \\
\midrule
\multirow{5}{*}{Video Colorization}
& ToonCrafter~\cite{xing2024tooncrafter} & 0.601 & 19.025 & 0.891 & 124.497 & 0.287 \\
& Anidoc~\cite{meng2024anidoc} & 0.518 & 18.341 & 0.836 & 158.382 & 0.463 \\
& LVCD~\cite{huang2024lvcd} & 0.556 & 15.944 & 0.865 & 138.828 & 0.454 \\
& ToonComposer~\cite{tooncomposer} & 0.581 & 18.082 & \textbf{0.929} & 132.319 & 0.304 \\
& \textbf{Ours} & \textbf{0.689} & \textbf{20.521} & 0.925 & \textbf{119.514} & \textbf{0.193} \\
\bottomrule
\end{tabular}
\end{adjustbox}
\end{table}

\subsection{Qualitative Results}


As shown in Figure~\ref{fig: comparison}, qualitative evaluations demonstrate that our method outperforms state-of-the-art baselines in both image and video sketch colorization, excelling in physical detail preservation, semantic consistency, and temporal stability. For image colorization, ColorizeDiffusion’s style reproduction is constrained by its training dataset and sensitive to hyperparameters, leading to unstable alignment; MangaNinja maintains basic style consistency but fails at semantic matching (e.g., mismatched object color assignments) and performs poorly with multiple references even with guidance; MagicColor achieves instance-level semantic consistency for single or multiple references but suffers from fine-grained color errors on small textures or edges, while our method seamlessly supports both single and multiple reference inputs, delivering fine-grained semantic consistency and preserving high-frequency details like fabric textures and hair strands. 
In video colorization, LVCD fails to track large-scale motion transformations between input sketches and references, resulting in disjointed color mappings; ToonCrafter struggles with multiple references, misinterpreting instance-specific color information and causing cross-frame color conflicts; Anidoc lacks robustness in complex motion scenarios (e.g., references with inherent motion or camera trajectories), leading to inter-frame misalignment, discoloration, and lost instance identity, whereas our method consistently preserves instance identity (e.g., stable object/character colors) and maintains smooth temporal coherence even in large-scale motion or multi-reference scenarios, avoiding flickering, color conflicts, and misalignment to deliver visually coherent outputs.

\section{Ablation Study}

\begin{figure}[t]
\centering
\includegraphics[width=\linewidth]{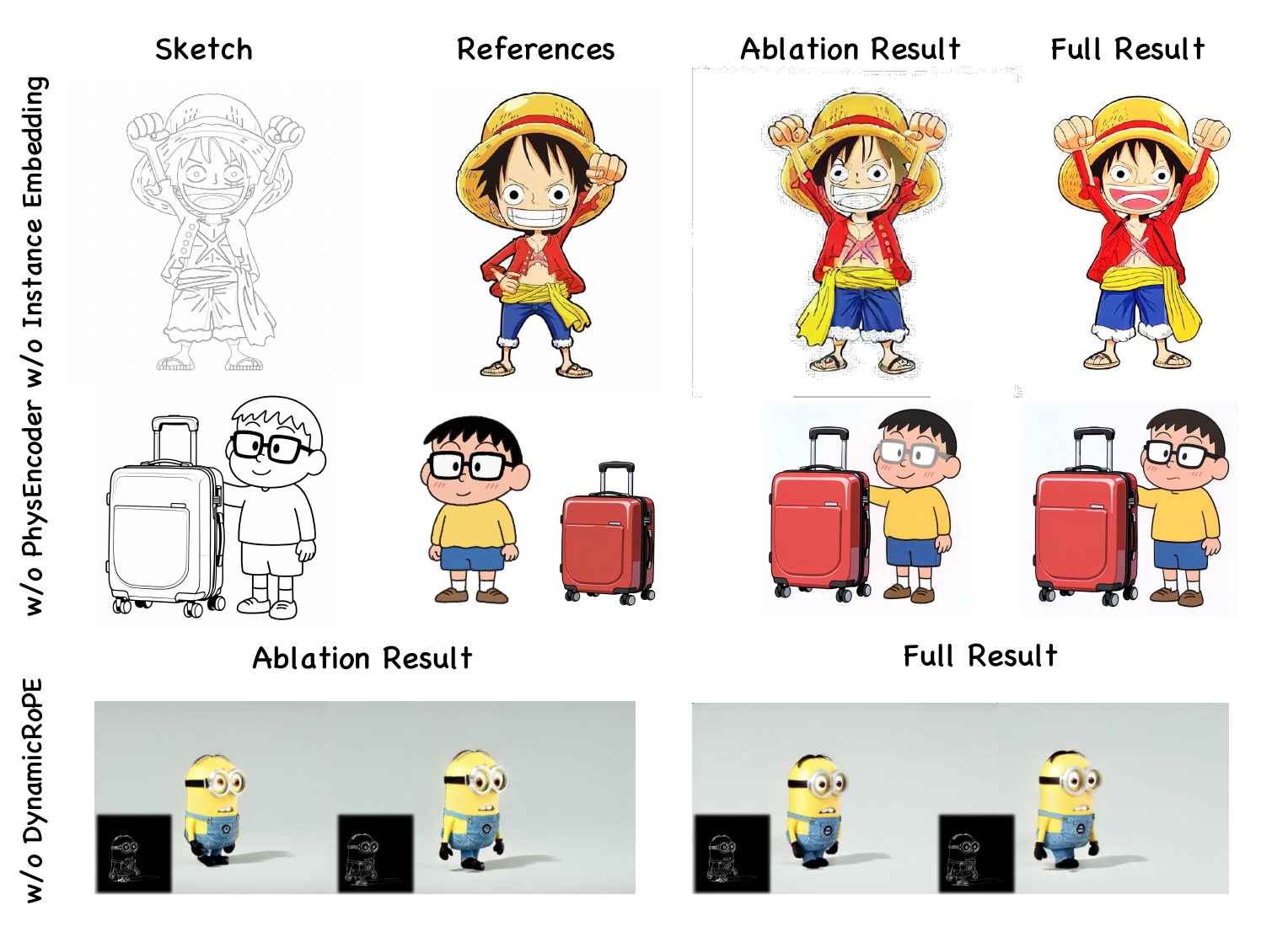}
\caption{\textbf{Ablation Result of each component.} After removing the Visual Reference Enhancement and PhysEncoder, the consistency of detail semantics decreases. When dynamic encoding is installed, the movement of the main subject becomes more natural.}
\label{fig: ablation}
\end{figure}

To investigate the contribution of key components in our framework, we conduct systematic ablation experiments by sequentially removing three core modules: Instance Embedding, Visual Reference Enhancement, and Sketch-based Dynamic RoPE. 
As shown in Table~\ref{tab: ablation}, the Full Model achieves the best overall performance across most metrics, confirming the effectiveness of our integrated design.

\begin{table}[t]
\centering
\caption{Detailed ablation analysis on image and video tasks.}
\label{tab: ablation}
\small
\begin{adjustbox}{width=\linewidth}
\begin{tabular}{@{}lccccc@{}}
\toprule
\textbf{Component} & \textbf{Task} & \textbf{SSIM $\uparrow$} & \textbf{LPIPS $\downarrow$} & \textbf{ FID$\downarrow$} & \textbf{ T-Consist$\uparrow$} \\
\midrule
\multirow{2}{*}{Full Model} 
& Image & \textbf{0.741} & \textbf{0.211} & \textbf{23.761} & - \\
& Video & \textbf{0.688} & \textbf{0.193} & \textbf{119.514} & \textbf{0.971} \\
\midrule
\multirow{2}{*}{w/o PhysEncoder}
& Image & 0.737 & 0.219 & 24.955 & - \\
& Video & 0.673 & 0.209 & 124.380 & 0.966 \\
\midrule
\multirow{2}{*}{w/o Instance Embedding}
& Image & 0.731 & 0.207 & 24.019 & - \\
& Video & 0.646 & 0.231 & 126.841 & 0.962 \\
\midrule
\multirow{2}{*}{w/o Dynamic RoPE}
& Image & - & - & - & - \\
& Video & 0.679 & 0.201 & 121.901 & 0.961 \\
\bottomrule
\end{tabular}
\end{adjustbox}
\end{table}

\noindent\textbf{Effectiveness of Visual Reference Fusion:} As shown in Fig.~\ref{fig: ablation} and Table~\ref{tab: ablation}, ablating this module leads to significant performance degradation: key metrics (e.g., FID, LPIPS) drop dramatically, and visual results suffer from severe identity inconsistency. For images, precise instance-color mapping is lost; for videos, cross-frame color confusion between reference instances becomes prominent. For example, the main subject’s face and feet in the first row of Fig.~\ref{fig: ablation} exhibit color overflow and mismatched instance coloring. This confirms the module’s critical role in maintaining consistent per-instance identity across both image and video colorization.

\noindent\textbf{Effectiveness of Physical Detail Reinforcement:} Removing this module degrades quality as quantified in Table~\ref{tab: ablation}: FID and LPIPS increase notably, especially for videos. Visually, images lose edge sharpness and texture granularity, while videos suffer from cross-frame detail inconsistency, particularly in environment shadows and object fine details. These results verify the module’s necessity for preserving physical detail fidelity in both tasks.

\noindent\textbf{Effectiveness of Dynamic RoPE:} While this ablation causes moderate drops in quantitative metrics, it induces distinct temporal consistency issues in videos. As shown in Fig.~\ref{fig: ablation}, without Dynamic RoPE, the subject exhibits obvious frame jitter and motion misalignment when walking, failing to maintain natural, fluent movement. For images, fine-grained details also blur slightly. This highlights that the module ensures smooth motion for videos and stabilizes spatial features for images.

\section{Conclusion}
In this paper, we propose Uni-Animator, a DiT-based unified framework for image and video sketch colorization, addressing key challenges of imprecise color transfer, insufficient detail preservation, and motion-induced temporal inconsistency. Equipped with visual reference enhancement, DINO-based physical detail reinforcement, and sketch-based dynamic RoPE, our method achieves competitive performance on both tasks, matching specialized methods while offering a flexible unified pipeline. Evaluations validate its superiority in detail fidelity, style alignment, and temporal coherence. Future work will focus on real-time processing and higher-resolution outputs.

{
    \small
    \bibliographystyle{plainnat}
    \bibliography{main}

@String(CVPR= {IEEE Conf. Comput. Vis. Pattern Recog.})

@String(TOG= {ACM Trans. Graph.})

@String(ICIP = {IEEE Int. Conf. Image Process.})

@String(ICLR = {Int. Conf. Learn. Represent.})

@String(AAAI = {AAAI})

@String(CVPR  = {CVPR})

@String(TOG   = {ACM TOG})

@String(ICIP  = {ICIP})

@String(ICLR  = {ICLR})

@misc{coloridiffusionv2,
      title={Enhancing Reference-based Sketch Colorization via Separating Reference Representations}, 
      author={Dingkun Yan and Xinrui Wang and Zhuoru Li and Suguru Saito and Yusuke Iwasawa and Yutaka Matsuo and Jiaxian Guo},
      year={2025},
      eprint={2508.17620},
      archivePrefix={arXiv},
      primaryClass={cs.GR},
      url={https://arxiv.org/abs/2508.17620}, 
}

@misc{Magic3DSketch,
      title={Magic3DSketch: Create Colorful 3D Models From Sketch-Based 3D Modeling Guided by Text and Language-Image Pre-Training}, 
      author={Ying Zang and Yidong Han and Chaotao Ding and Jianqi Zhang and Tianrun Chen},
      year={2024},
      eprint={2407.19225},
      archivePrefix={arXiv},
      primaryClass={cs.CV},
      url={https://arxiv.org/abs/2407.19225}, 
}

@misc{AnimeColor,
      title={AnimeColor: Reference-based Animation Colorization with Diffusion Transformers}, 
      author={Yuhong Zhang and Liyao Wang and Han Wang and Danni Wu and Zuzeng Lin and Feng Wang and Li Song},
      year={2025},
      eprint={2507.20158},
      archivePrefix={arXiv},
      primaryClass={cs.CV},
      url={https://arxiv.org/abs/2507.20158}, 
}

@misc{LayerAnimate,
      title={LayerAnimate: Layer-level Control for Animation}, 
      author={Yuxue Yang and Lue Fan and Zuzeng Lin and Feng Wang and Zhaoxiang Zhang},
      year={2025},
      eprint={2501.08295},
      archivePrefix={arXiv},
      primaryClass={cs.CV},
      url={https://arxiv.org/abs/2501.08295}, 
}

@misc{sam,
      title={Segment Anything}, 
      author={Alexander Kirillov and Eric Mintun and Nikhila Ravi and Hanzi Mao and Chloe Rolland and Laura Gustafson and Tete Xiao and Spencer Whitehead and Alexander C. Berg and Wan-Yen Lo and Piotr Dollár and Ross Girshick},
      year={2023},
      eprint={2304.02643},
      archivePrefix={arXiv},
      primaryClass={cs.CV},
      url={https://arxiv.org/abs/2304.02643}, 
}

@misc{sakuga,
      title={Sakuga-42M Dataset: Scaling Up Cartoon Research}, 
      author={Zhenglin Pan},
      year={2024},
      eprint={2405.07425},
      archivePrefix={arXiv},
      primaryClass={cs.CV},
      url={https://arxiv.org/abs/2405.07425}, 
}

@article{LayerT2V,
  title={LayerT2V: Interactive Multi-Object Trajectory Layering for Video Generation},
  author={Cen, Kangrui and Zhao, Baixuan and Xin, Yi and Luo, Siqi and Zhai, Guangtao and Liu, Xiaohong},
  journal={arXiv preprint arXiv:2508.04228},
  year={2025}
}

@misc{followyourcolor,
      title={Follow-Your-Color: Multi-Instance Sketch Colorization}, 
      author={Yinhan Zhang and Yue Ma and Bingyuan Wang and Qifeng Chen and Zeyu Wang},
      year={2025},
      eprint={2503.16948},
      archivePrefix={arXiv},
      primaryClass={cs.CV},
      url={https://arxiv.org/abs/2503.16948}, 
}

@misc{tooncomposer,
      title={ToonComposer: Streamlining Cartoon Production with Generative Post-Keyframing}, 
      author={Lingen Li and Guangzhi Wang and Zhaoyang Zhang and Yaowei Li and Xiaoyu Li and Qi Dou and Jinwei Gu and Tianfan Xue and Ying Shan},
      year={2025},
      eprint={2508.10881},
      archivePrefix={arXiv},
      primaryClass={cs.CV},
      url={https://arxiv.org/abs/2508.10881}, 
}

@article{xing2024tooncrafter,
  title={ToonCrafter: Generative Cartoon Interpolation},
  author={Xing, Jinbo and Liu, Hanyuan and Xia, Menghan and Zhang, Yong and Wang, Xintao and Shan, Ying and Wong, Tien-Tsin},
  journal={arXiv preprint arXiv:2405.12345},
  year={2024}
}

@article{dit,
  title={Scalable Diffusion Models with Transformers},
  author={Peebles, William and Xie, Saining},
  journal={arXiv preprint arXiv:2303.12345},
  year={2023}
}

@article{CGAN,
  title={Image-to-Image Translation with Conditional Adversarial Networks},
  author={Isola, Phillip and Zhu, Jun-Yan and Zhou, Tinghui and Efros, Alexei A.},
  journal={arXiv preprint arXiv:1611.07004},
  year={2017}
}

@article{ho2020denoising,
  title={Denoising Diffusion Probabilistic Models},
  author={Ho, Jonathan and Jain, Ajay and Abbeel, Pieter},
  journal={arXiv preprint arXiv:2006.11239},
  year={2020}
}

@article{controlnet,
  title={Adding Conditional Control to Text-to-Image Diffusion Models},
  author={Zhang, Lvmin and Agrawala, Maneesh},
  journal={arXiv preprint arXiv:2302.05543},
  year={2023}
}

@article{meng2024anidoc,
  title={Anidoc: Animation creation made easier},
  author={Meng, Yihao and Ouyang, Hao and Wang, Hanlin and Wang, Qiuyu and Wang, Wen and Cheng, Ka Leong and Liu, Zhiheng and Shen, Yujun and Qu, Huamin},
  journal={arXiv preprint arXiv:2412.14173},
  year={2024}
}

@article{huang2024lvcd,
  title={Lvcd: reference-based lineart video colorization with diffusion models},
  author={Huang, Zhitong and Zhang, Mohan and Liao, Jing},
  journal={ACM Transactions on Graphics (TOG)},
  volume={43},
  number={6},
  pages={1--11},
  year={2024},
  publisher={ACM New York, NY, USA}
}

@article{wang2025wan,
  title={Wan: Open and advanced large-scale video generative models},
  author={Wang, Ang and Ai, Baole and Wen, Bin and Mao, Chaojie and Xie, Chen-Wei and Chen, Di and Yu, Feiwu and Zhao, Haiming and Yang, Jianxiao and Zeng, Jianyuan and others},
  journal={arXiv preprint arXiv:2503.20314},
  year={2025}
}

@article{jiang2024exploring,
  title={Exploring the Frontiers of Animation Video Generation in the Sora Era: Method, Dataset and Benchmark},
  author={Jiang, Yudong and Xu, Baohan and Yang, Siqian and Yin, Mingyu and Liu, Jing and Xu, Chao and Wang, Siqi and Wu, Yidi and Zhu, Bingwen and Xu, Jixuan and others},
  journal={arXiv preprint arXiv:2412.10255},
  year={2024}
}

@article{ho2022video,
  title={Video diffusion models},
  author={Ho, Jonathan and Salimans, Tim and Gritsenko, Alexey and Chan, William and Norouzi, Mohammad and Fleet, David J},
  journal={Advances in Neural Information Processing Systems},
  volume={35},
  pages={8633--8646},
  year={2022}
}

@inproceedings{Dynamicrafter,
  title={Dynamicrafter: Animating open-domain images with video diffusion priors},
  author={Xing, Jinbo and Xia, Menghan and Zhang, Yong and Chen, Haoxin and Yu, Wangbo and Liu, Hanyuan and Liu, Gongye and Wang, Xintao and Shan, Ying and Wong, Tien-Tsin},
  booktitle={European Conference on Computer Vision},
  pages={399--417},
  year={2024},
  organization={Springer}
}

@article{hu2022lora,
  title={Lora: Low-rank adaptation of large language models.},
  author={Hu, Edward J and Shen, Yelong and Wallis, Phillip and Allen-Zhu, Zeyuan and Li, Yuanzhi and Wang, Shean and Wang, Lu and Chen, Weizhu and others},
  journal={ICLR},
  volume={1},
  number={2},
  pages={3},
  year={2022}
}

@inproceedings{sketch-to-photo,
  title={Adversarial open domain adaptation for sketch-to-photo synthesis},
  author={Xiang, Xiaoyu and Liu, Ding and Yang, Xiao and Zhu, Yiheng and Shen, Xiaohui and Allebach, Jan P},
  booktitle={Proceedings of the IEEE/CVF Winter Conference on Applications of Computer Vision},
  pages={1434--1444},
  year={2022}
}

@inproceedings{lapips,
  title={The unreasonable effectiveness of deep features as a perceptual metric},
  author={Zhang, Richard and Isola, Phillip and Efros, Alexei A and Shechtman, Eli and Wang, Oliver},
  booktitle={Proceedings of the IEEE conference on computer vision and pattern recognition},
  pages={586--595},
  year={2018}
}

@article{ssim,
  title={Image quality assessment: Unifying structure and texture similarity},
  author={Ding, Keyan and Ma, Kede and Wang, Shiqi and Simoncelli, Eero P},
  journal={IEEE transactions on pattern analysis and machine intelligence},
  volume={44},
  number={5},
  pages={2567--2581},
  year={2020},
  publisher={IEEE}
}

@article{ColorFlow,
  title={ColorFlow: Retrieval-Augmented Image Sequence Colorization},
  author={Zhuang, Junhao and Ju, Xuan and Zhang, Zhaoyang and Liu, Yong and Zhang, Shiyi and Yuan, Chun and Shan, Ying},
  journal={arXiv preprint arXiv:2412.11815},
  year={2024}
}

@article{Cobra,
  title={Cobra: Efficient Line Art COlorization with BRoAder References},
  author={Zhuang, Junhao and Li, Lingen and Ju, Xuan and Zhang, Zhaoyang and Yuan, Chun and Shan, Ying},
  journal={arXiv preprint arXiv:2504.12240},
  year={2025}
}

@article{ye2023ip,
  title={Ip-adapter: Text compatible image prompt adapter for text-to-image diffusion models},
  author={Ye, Hu and Zhang, Jun and Liu, Sibo and Han, Xiao and Yang, Wei},
  journal={arXiv preprint arXiv:2308.06721},
  year={2023}
}

@inproceedings{zhu2024champ,
  title={Champ: Controllable and consistent human image animation with 3d parametric guidance},
  author={Zhu, Shenhao and Chen, Junming Leo and Dai, Zuozhuo and Dong, Zilong and Xu, Yinghui and Cao, Xun and Yao, Yao and Zhu, Hao and Zhu, Siyu},
  booktitle={European Conference on Computer Vision},
  pages={145--162},
  year={2024},
  organization={Springer}
}

@inproceedings{hu2024animate,
  title={Animate anyone: Consistent and controllable image-to-video synthesis for character animation},
  author={Hu, Li},
  booktitle={Proceedings of the IEEE/CVF Conference on Computer Vision and Pattern Recognition},
  pages={8153--8163},
  year={2024}
}

@article{xing2024make,
  title={Make-your-video: Customized video generation using textual and structural guidance},
  author={Xing, Jinbo and Xia, Menghan and Liu, Yuxin and Zhang, Yuechen and Zhang, Yong and He, Yingqing and Liu, Hanyuan and Chen, Haoxin and Cun, Xiaodong and Wang, Xintao and others},
  journal={IEEE Transactions on Visualization and Computer Graphics},
  year={2024},
  publisher={IEEE}
}

@article{wang2025cinemaster,
  title={CineMaster: A 3D-Aware and Controllable Framework for Cinematic Text-to-Video Generation},
  author={Wang, Qinghe and Luo, Yawen and Shi, Xiaoyu and Jia, Xu and Lu, Huchuan and Xue, Tianfan and Wang, Xintao and Wan, Pengfei and Zhang, Di and Gai, Kun},
  journal={arXiv preprint arXiv:2502.08639},
  year={2025}
}

@inproceedings{blattmann2023align,
  title={Align your latents: High-resolution video synthesis with latent diffusion models},
  author={Blattmann, Andreas and Rombach, Robin and Ling, Huan and Dockhorn, Tim and Kim, Seung Wook and Fidler, Sanja and Kreis, Karsten},
  booktitle={Proceedings of the IEEE/CVF Conference on Computer Vision and Pattern Recognition},
  pages={22563--22575},
  year={2023}
}

@article{blattmann2023stable,
  title={Stable video diffusion: Scaling latent video diffusion models to large datasets},
  author={Blattmann, Andreas and Dockhorn, Tim and Kulal, Sumith and Mendelevitch, Daniel and Kilian, Maciej and Lorenz, Dominik and Levi, Yam and English, Zion and Voleti, Vikram and Letts, Adam and others},
  journal={arXiv preprint arXiv:2311.15127},
  year={2023}
}

@article{kong2024hunyuanvideo,
  title={Hunyuanvideo: A systematic framework for large video generative models},
  author={Kong, Weijie and Tian, Qi and Zhang, Zijian and Min, Rox and Dai, Zuozhuo and Zhou, Jin and Xiong, Jiangfeng and Li, Xin and Wu, Bo and Zhang, Jianwei and others},
  journal={arXiv preprint arXiv:2412.03603},
  year={2024}
}

@article{yang2024cogvideox,
  title={Cogvideox: Text-to-video diffusion models with an expert transformer},
  author={Yang, Zhuoyi and Teng, Jiayan and Zheng, Wendi and Ding, Ming and Huang, Shiyu and Xu, Jiazheng and Yang, Yuanming and Hong, Wenyi and Zhang, Xiaohan and Feng, Guanyu and others},
  journal={arXiv preprint arXiv:2408.06072},
  year={2024}
}

@article{su2024roformer,
  title={Roformer: Enhanced transformer with rotary position embedding},
  author={Su, Jianlin and Ahmed, Murtadha and Lu, Yu and Pan, Shengfeng and Bo, Wen and Liu, Yunfeng},
  journal={Neurocomputing},
  volume={568},
  pages={127063},
  year={2024},
  publisher={Elsevier}
}

@article{loshchilov2017decoupled,
  title={Decoupled weight decay regularization},
  author={Loshchilov, Ilya and Hutter, Frank},
  journal={arXiv preprint arXiv:1711.05101},
  year={2017}
}

@inproceedings{videointerpolation,
  title={Enhanced deep animation video interpolation},
  author={Shen, Wang and Ming, Cheng and Bao, Wenbo and Zhai, Guangtao and Chenn, Li and Gao, Zhiyong},
  booktitle={2022 IEEE International Conference on Image Processing (ICIP)},
  pages={31--35},
  year={2022},
  organization={IEEE}
}

@article{liu2025manganinja,
  title={MangaNinja: Line Art Colorization with Precise Reference Following},
  author={Liu, Zhiheng and Cheng, Ka Leong and Chen, Xi and Xiao, Jie and Ouyang, Hao and Zhu, Kai and Liu, Yu and Shen, Yujun and Chen, Qifeng and Luo, Ping},
  journal={arXiv preprint arXiv:2501.08332},
  year={2025}
}

@InProceedings{style2paint,
  author={Lvmin Zhang and Chengze Li and Edgar Simo-Serra and Yi Ji and Tien-Tsin Wong and Chunping Liu}, 
  booktitle={IEEE/CVF Conference on Computer Vision and Pattern Recognition (CVPR)}, 
  title={User-Guided Line Art Flat Filling with Split Filling Mechanism}, 
  year={2021}, 
}

@misc{AnimeDiffusion,
      title={AnimeDiffusion: Anime Face Line Drawing Colorization via Diffusion Models}, 
      author={Yu Cao and Xiangqiao Meng and P. Y. Mok and Xueting Liu and Tong-Yee Lee and Ping Li},
      year={2023},
      eprint={2303.11137},
      archivePrefix={arXiv},
      primaryClass={cs.CV},
      url={https://arxiv.org/abs/2303.11137}, 
}

@misc{SGA,
      title={Eliminating Gradient Conflict in Reference-based Line-Art Colorization}, 
      author={Zekun Li and Zhengyang Geng and Zhao Kang and Wenyu Chen and Yibo Yang},
      year={2022},
      eprint={2207.06095},
      archivePrefix={arXiv},
      primaryClass={cs.CV},
      url={https://arxiv.org/abs/2207.06095}, 
}

@misc{ChromaGAN,
      title={ChromaGAN: Adversarial Picture Colorization with Semantic Class Distribution}, 
      author={Patricia Vitoria and Lara Raad and Coloma Ballester},
      year={2020},
      eprint={1907.09837},
      archivePrefix={arXiv},
      primaryClass={cs.CV},
      url={https://arxiv.org/abs/1907.09837}, 
}

@misc{pix2pix,
      title={Image-to-Image Translation with Conditional Adversarial Networks}, 
      author={Phillip Isola and Jun-Yan Zhu and Tinghui Zhou and Alexei A. Efros},
      year={2018},
      eprint={1611.07004},
      archivePrefix={arXiv},
      primaryClass={cs.CV},
      url={https://arxiv.org/abs/1611.07004}, 
}

@misc{BigColor,
      title={BigColor: Colorization using a Generative Color Prior for Natural Images}, 
      author={Geonung Kim and Kyoungkook Kang and Seongtae Kim and Hwayoon Lee and Sehoon Kim and Jonghyun Kim and Seung-Hwan Baek and Sunghyun Cho},
      year={2022},
      eprint={2207.09685},
      archivePrefix={arXiv},
      primaryClass={cs.CV},
      url={https://arxiv.org/abs/2207.09685}, 
}

@article{VCGAN,
   title={VCGAN: Video Colorization With Hybrid Generative Adversarial Network},
   volume={25},
   ISSN={1941-0077},
   url={http://dx.doi.org/10.1109/TMM.2022.3154600},
   DOI={10.1109/tmm.2022.3154600},
   journal={IEEE Transactions on Multimedia},
   publisher={Institute of Electrical and Electronics Engineers (IEEE)},
   author={Zhao, Yuzhi and Po, Lai-Man and Yu, Wing-Yin and Rehman, Yasar Abbas Ur and Liu, Mengyang and Zhang, Yujia and Ou, Weifeng},
   year={2023},
   pages={3017–3032} }

@misc{DINOv2,
      title={DINOv2: Learning Robust Visual Features without Supervision}, 
      author={Maxime Oquab and Timothée Darcet and Théo Moutakanni and Huy Vo and Marc Szafraniec and Vasil Khalidov and Pierre Fernandez and Daniel Haziza and Francisco Massa and Alaaeldin El-Nouby and Mahmoud Assran and Nicolas Ballas and Wojciech Galuba and Russell Howes and Po-Yao Huang and Shang-Wen Li and Ishan Misra and Michael Rabbat and Vasu Sharma and Gabriel Synnaeve and Hu Xu and Hervé Jegou and Julien Mairal and Patrick Labatut and Armand Joulin and Piotr Bojanowski},
      year={2024},
      eprint={2304.07193},
      archivePrefix={arXiv},
      primaryClass={cs.CV},
      url={https://arxiv.org/abs/2304.07193}, 
}

@misc{t5,
      title={Exploring the Limits of Transfer Learning with a Unified Text-to-Text Transformer}, 
      author={Colin Raffel and Noam Shazeer and Adam Roberts and Katherine Lee and Sharan Narang and Michael Matena and Yanqi Zhou and Wei Li and Peter J. Liu},
      year={2023},
      eprint={1910.10683},
      archivePrefix={arXiv},
      primaryClass={cs.LG},
      url={https://arxiv.org/abs/1910.10683}, 
}

@misc{Wan,
      title={Wan: Open and Advanced Large-Scale Video Generative Models}, 
      author={Team Wan and Ang Wang and Baole Ai and Bin Wen and Chaojie Mao and Chen-Wei Xie and Di Chen and Feiwu Yu and Haiming Zhao and Jianxiao Yang and Jianyuan Zeng and Jiayu Wang and Jingfeng Zhang and Jingren Zhou and Jinkai Wang and Jixuan Chen and Kai Zhu and Kang Zhao and Keyu Yan and Lianghua Huang and Mengyang Feng and Ningyi Zhang and Pandeng Li and Pingyu Wu and Ruihang Chu and Ruili Feng and Shiwei Zhang and Siyang Sun and Tao Fang and Tianxing Wang and Tianyi Gui and Tingyu Weng and Tong Shen and Wei Lin and Wei Wang and Wei Wang and Wenmeng Zhou and Wente Wang and Wenting Shen and Wenyuan Yu and Xianzhong Shi and Xiaoming Huang and Xin Xu and Yan Kou and Yangyu Lv and Yifei Li and Yijing Liu and Yiming Wang and Yingya Zhang and Yitong Huang and Yong Li and You Wu and Yu Liu and Yulin Pan and Yun Zheng and Yuntao Hong and Yupeng Shi and Yutong Feng and Zeyinzi Jiang and Zhen Han and Zhi-Fan Wu and Ziyu Liu},
      year={2025},
      eprint={2503.20314},
      archivePrefix={arXiv},
      primaryClass={cs.CV},
      url={https://arxiv.org/abs/2503.20314}, 
}

@misc{RAFT,
      title={RAFT: Recurrent All-Pairs Field Transforms for Optical Flow}, 
      author={Zachary Teed and Jia Deng},
      year={2020},
      eprint={2003.12039},
      archivePrefix={arXiv},
      primaryClass={cs.CV},
      url={https://arxiv.org/abs/2003.12039}, 
}

@inproceedings{ma2024follow,
  title={Follow your pose: Pose-guided text-to-video generation using pose-free videos},
  author={Ma, Yue and He, Yingqing and Cun, Xiaodong and Wang, Xintao and Chen, Siran and Li, Xiu and Chen, Qifeng},
  booktitle={Proceedings of the AAAI Conference on Artificial Intelligence},
  volume={38},
  number={5},
  pages={4117--4125},
  year={2024}
}

@inproceedings{ma2024followpose,
  title={Follow your pose: Pose-guided text-to-video generation using pose-free videos},
  author={Ma, Yue and He, Yingqing and Cun, Xiaodong and Wang, Xintao and Chen, Siran and Li, Xiu and Chen, Qifeng},
  booktitle={Proceedings of the AAAI Conference on Artificial Intelligence},
  volume={38},
  number={5},
  pages={4117--4125},
  year={2024}
}

@article{ma2025followcreation,
  title={Follow-Your-Creation: Empowering 4D Creation through Video Inpainting},
  author={Ma, Yue and Feng, Kunyu and Zhang, Xinhua and Liu, Hongyu and Zhang, David Junhao and Xing, Jinbo and Zhang, Yinhan and Yang, Ayden and Wang, Zeyu and Chen, Qifeng},
  journal={arXiv preprint arXiv:2506.04590},
  year={2025}
}

@article{ma2026fastvmt,
  title={FastVMT: Eliminating Redundancy in Video Motion Transfer},
  author={Ma, Yue and Wang, Zhikai and Ren, Tianhao and Zheng, Mingzhe and Liu, Hongyu and Guo, Jiayi and Fong, Mark and Xue, Yuxuan and Zhao, Zixiang and Schindler, Konrad and others},
  journal={arXiv preprint arXiv:2602.05551},
  year={2026}
}

@article{ma2025followyourmotion,
  title={Follow-Your-Motion: Video Motion Transfer via Efficient Spatial-Temporal Decoupled Finetuning},
  author={Ma, Yue and Liu, Yulong and Zhu, Qiyuan and Yang, Ayden and Feng, Kunyu and Zhang, Xinhua and Li, Zhifeng and Han, Sirui and Qi, Chenyang and Chen, Qifeng},
  journal={arXiv preprint arXiv:2506.05207},
  year={2025}
}

@article{ma2025followfaster,
  title={Follow-your-emoji-faster: Towards efficient, fine-controllable, and expressive freestyle portrait animation},
  author={Ma, Yue and Yan, Zexuan and Liu, Hongyu and Wang, Hongfa and Pan, Heng and He, Yingqing and Yuan, Junkun and Zeng, Ailing and Cai, Chengfei and Shum, Heung-Yeung and others},
  journal={arXiv preprint arXiv:2509.16630},
  year={2025}
}

@article{ma2025controllable,
  title={Controllable Video Generation: A Survey},
  author={Ma, Yue and Feng, Kunyu and Hu, Zhongyuan and Wang, Xinyu and Wang, Yucheng and Zheng, Mingzhe and He, Xuanhua and Zhu, Chenyang and Liu, Hongyu and He, Yingqing and others},
  journal={arXiv preprint arXiv:2507.16869},
  year={2025}
}

@inproceedings{ma2024followyouremoji,
  title={Follow-your-emoji: Fine-controllable and expressive freestyle portrait animation},
  author={Ma, Yue and Liu, Hongyu and Wang, Hongfa and Pan, Heng and He, Yingqing and Yuan, Junkun and Zeng, Ailing and Cai, Chengfei and Shum, Heung-Yeung and Liu, Wei and others},
  booktitle={SIGGRAPH Asia 2024 Conference Papers},
  pages={1--12},
  year={2024}
}

@inproceedings{ma2025followyourclick,
  title={Follow-Your-Click: Open-domain Regional Image Animation via Motion Prompts},
  author={Ma, Yue and He, Yingqing and Wang, Hongfa and Wang, Andong and Shen, Leqi and Qi, Chenyang and Ying, Jixuan and Cai, Chengfei and Li, Zhifeng and Shum, Heung-Yeung and others},
  booktitle={Proceedings of the AAAI Conference on Artificial Intelligence},
  volume={39},
  number={6},
  pages={6018--6026},
  year={2025}
}

@article{long2025follow,
  title={Follow-your-shape: Shape-aware image editing via trajectory-guided region control},
  author={Long, Zeqian and Zheng, Mingzhe and Feng, Kunyu and Zhang, Xinhua and Liu, Hongyu and Yang, Harry and Zhang, Linfeng and Chen, Qifeng and Ma, Yue},
  journal={arXiv preprint arXiv:2508.08134},
  year={2025}
}

@inproceedings{
z1,
title={{KABB}: Knowledge-Aware Bayesian Bandits for Dynamic Expert Coordination in Multi-Agent Systems},
author={Jusheng Zhang and Zimeng Huang and Yijia Fan and Ningyuan Liu and Mingyan Li and Zhuojie Yang and Jiawei Yao and Jian Wang and Keze Wang},
booktitle={Forty-second International Conference on Machine Learning},
year={2025},
url={https://openreview.net/forum?id=AKvy9a4jho}
}

@inproceedings{
z2,
title={{GAM}-Agent: Game-Theoretic and Uncertainty-Aware Collaboration for Complex Visual Reasoning},
author={Jusheng Zhang and Yijia Fan and Wenjun Lin and Ruiqi Chen and Haoyi Jiang and Wenhao Chai and Jian Wang and Keze Wang},
booktitle={The Thirty-ninth Annual Conference on Neural Information Processing Systems},
year={2025},
url={https://openreview.net/forum?id=EKJhU5ioSo}
}

@misc{z3,
      title={CF-VLM:CounterFactual Vision-Language Fine-tuning}, 
      author={Jusheng Zhang and Kaitong Cai and Yijia Fan and Jian Wang and Keze Wang},
      year={2025},
      eprint={2506.17267},
      archivePrefix={arXiv},
      primaryClass={cs.LG},
      url={https://arxiv.org/abs/2506.17267}, 
}

@inproceedings{
z4,
title={{MAT}-Agent: Adaptive Multi-Agent Training Optimization},
author={Jusheng Zhang and Kaitong Cai and Yijia Fan and Ningyuan Liu and Keze Wang},
booktitle={The Thirty-ninth Annual Conference on Neural Information Processing Systems},
year={2025},
url={https://openreview.net/forum?id=YDWRTYgR79}
}

@article{zhang2025easycontrol,
  title={Easycontrol: Adding efficient and flexible control for diffusion transformer},
  author={Zhang, Yuxuan and Yuan, Yirui and Song, Yiren and Wang, Haofan and Liu, Jiaming},
  journal={arXiv preprint arXiv:2503.07027},
  year={2025}
}

@inproceedings{zhang2024ssr,
  title={Ssr-encoder: Encoding selective subject representation for subject-driven generation},
  author={Zhang, Yuxuan and Song, Yiren and Liu, Jiaming and Wang, Rui and Yu, Jinpeng and Tang, Hao and Li, Huaxia and Tang, Xu and Hu, Yao and Pan, Han and others},
  booktitle={Proceedings of the IEEE/CVF Conference on Computer Vision and Pattern Recognition},
  pages={8069--8078},
  year={2024}
}

@article{song2025layertracer,
  title={LayerTracer: Cognitive-Aligned Layered SVG Synthesis via Diffusion Transformer},
  author={Song, Yiren and Chen, Danze and Shou, Mike Zheng},
  journal={arXiv preprint arXiv:2502.01105},
  year={2025}
}

@article{song2025makeanything,
  title={MakeAnything: Harnessing Diffusion Transformers for Multi-Domain Procedural Sequence Generation},
  author={Song, Yiren and Liu, Cheng and Shou, Mike Zheng},
  journal={arXiv preprint arXiv:2502.01572},
  year={2025}
}

@article{huang2025photodoodle,
  title={Photodoodle: Learning artistic image editing from few-shot pairwise data},
  author={Huang, Shijie and Song, Yiren and Zhang, Yuxuan and Guo, Hailong and Wang, Xueyin and Shou, Mike Zheng and Liu, Jiaming},
  journal={arXiv preprint arXiv:2502.14397},
  year={2025}
}

@article{wang2024taming,
  title={Taming rectified flow for inversion and editing},
  author={Wang, Jiangshan and Pu, Junfu and Qi, Zhongang and Guo, Jiayi and Ma, Yue and Huang, Nisha and Chen, Yuxin and Li, Xiu and Shan, Ying},
  journal={arXiv preprint arXiv:2411.04746},
  year={2024}
}

@inproceedings{feng2025dit4edit,
  title={Dit4edit: Diffusion transformer for image editing},
  author={Feng, Kunyu and Ma, Yue and Wang, Bingyuan and Qi, Chenyang and Chen, Haozhe and Chen, Qifeng and Wang, Zeyu},
  booktitle={Proceedings of the AAAI Conference on Artificial Intelligence},
  volume={39},
  number={3},
  pages={2969--2977},
  year={2025}
}

@article{wang2024cove,
  title={Cove: Unleashing the diffusion feature correspondence for consistent video editing},
  author={Wang, Jiangshan and Ma, Yue and Guo, Jiayi and Xiao, Yicheng and Huang, Gao and Li, Xiu},
  journal={Advances in Neural Information Processing Systems},
  volume={37},
  pages={96541--96565},
  year={2024}
}

@inproceedings{chen2025infinite,
  title={Infinite-Canvas: Higher-Resolution Video Outpainting with Extensive Content Generation},
  author={Chen, Qihua and Ma, Yue and Wang, Hongfa and Yuan, Junkun and Zhao, Wenzhe and Tian, Qi and Wang, Hongmei and Min, Shaobo and Chen, Qifeng and Liu, Wei},
  booktitle={Proceedings of the AAAI Conference on Artificial Intelligence},
  volume={39},
  number={2},
  pages={2150--2158},
  year={2025}
}

@inproceedings{zhu2025multibooth,
  title={Multibooth: Towards generating all your concepts in an image from text},
  author={Zhu, Chenyang and Li, Kai and Ma, Yue and He, Chunming and Li, Xiu},
  booktitle={Proceedings of the AAAI Conference on Artificial Intelligence},
  volume={39},
  number={10},
  pages={10923--10931},
  year={2025}
}

@article{zhu2024instantswap,
  title={Instantswap: Fast customized concept swapping across sharp shape differences},
  author={Zhu, Chenyang and Li, Kai and Ma, Yue and Tang, Longxiang and Fang, Chengyu and Chen, Chubin and Chen, Qifeng and Li, Xiu},
  journal={arXiv preprint arXiv:2412.01197},
  year={2024}
}

@inproceedings{yan2025eedit,
  title={Eedit: Rethinking the spatial and temporal redundancy for efficient image editing},
  author={Yan, Zexuan and Ma, Yue and Zou, Chang and Chen, Wenteng and Chen, Qifeng and Zhang, Linfeng},
  booktitle={Proceedings of the IEEE/CVF International Conference on Computer Vision},
  pages={17474--17484},
  year={2025}
}

@inproceedings{zhang2025magiccolor,
  title={Magiccolor: Multi-instance sketch colorization},
  author={Zhang, Yinhan and Ma, Yue and Wang, Bingyuan and Chen, Qifeng and Wang, Zeyu},
  booktitle={Proceedings of the IEEE/CVF International Conference on Computer Vision},
  pages={15205--15217},
  year={2025}
}

@inproceedings{liu2025avatarartist,
  title={Avatarartist: Open-domain 4d avatarization},
  author={Liu, Hongyu and Wang, Xuan and Wan, Ziyu and Ma, Yue and Chen, Jingye and Fan, Yanbo and Shen, Yujun and Song, Yibing and Chen, Qifeng},
  booktitle={Proceedings of the Computer Vision and Pattern Recognition Conference},
  pages={10758--10769},
  year={2025}
}

@article{liu2025multimotion,
  title={MultiMotion: Multi Subject Video Motion Transfer via Video Diffusion Transformer},
  author={Liu, Penghui and Wang, Jiangshan and Shen, Yutong and Mo, Shanhui and Qi, Chenyang and Ma, Yue},
  journal={arXiv preprint arXiv:2512.07500},
  year={2025}
}
}

\end{document}